\documentclass[10pt,conference]{IEEEtran}
\IEEEoverridecommandlockouts
\usepackage{cite}
\usepackage{amsmath,amssymb,amsfonts}
\usepackage{algorithmic}
\usepackage{graphicx}
\usepackage{textcomp}
\def\BibTeX{{\rm B\kern-.05em{\sc i\kern-.025em b}\kern-.08em
    T\kern-.1667em\lower.7ex\hbox{E}\kern-.125emX}}
\usepackage{pifont}
\newcommand{\cmark}{\ding{51}}%
\newcommand{\xmark}{\ding{55}}%
\usepackage{listings}
\usepackage{xcolor}
\usepackage{colortbl}
\usepackage{xparse}
\usepackage{multirow}
\usepackage{booktabs}
\usepackage{hyperref}

\definecolor{codegreen}{rgb}{0,0.6,0}
\definecolor{codegray}{rgb}{0.75,0.75,0.75}
\definecolor{codepurple}{rgb}{0.58,0,0.82}
\definecolor{backcolour}{rgb}{0.95,0.95,0.92}

\NewDocumentCommand{\codeword}{v}{%
\texttt{\textcolor{black}{#1}}%
}

\lstdefinestyle{mystyle}{
    backgroundcolor=\color{backcolour},   
    commentstyle=\color{codegreen},
    keywordstyle=\color{magenta},
    numberstyle=\tiny\color{codegray},
    stringstyle=\color{codepurple},
    basicstyle=\ttfamily\footnotesize,
    breakatwhitespace=false,         
    breaklines=true,                 
    captionpos=b,                    
    keepspaces=true,                 
    numbers=left,                    
    numbersep=5pt,                  
    showspaces=false,                
    showstringspaces=false,
    showtabs=false,                  
    tabsize=2
}

\begin{document}

\title{GraphGallery: A Platform for Fast Benchmarking and Easy Development of Graph Neural Networks Based Intelligent Software}

\author{\IEEEauthorblockN{Jintang Li, Kun Xu, Liang Chen*, Zibin Zheng}
\IEEEauthorblockA{\textit{Sun Yat-sen University, China} \\
{\{lijt55, xukun6\}@mail2.sysu.edu.cn}\\{\{chenliang6, zhzibin\}@mail.sysu.edu.cn}}
\and
\IEEEauthorblockN{Xiao Liu}
\IEEEauthorblockA{\textit{Deakin University, Australia} \\
    xiao.liu@deakin.edu.au}
}

\maketitle

\begin{abstract}

    Graph Neural Networks (GNNs) have recently shown to be powerful tools for representing and analyzing graph data. So far GNNs is becoming an increasingly critical role in software engineering including program analysis, type inference, and code representation.
    In this paper, we introduce GraphGallery, a platform for fast benchmarking and easy development of GNNs based software. GraphGallery is an easy-to-use platform that allows developers to automatically deploy GNNs even with less domain-specific knowledge. It offers a set of implementations of common GNN models based on mainstream deep learning frameworks. In addition, existing GNNs toolboxes such as PyG and DGL can be easily incorporated into the platform. Experiments demonstrate the reliability of implementations and superiority in fast coding. The official source code of GraphGallery is available at \url{https://github.com/EdisonLeeeee/GraphGallery} and a demo video can be found at \url{https://youtu.be/mv7Zs1YeaYo}.

\end{abstract}




\section{Introduction}
Graph Neural Networks (GNNs) have received a considerable amount of attention from academia and industries mainly due to the powerful ability in representing and modeling the relationships between nodes or edges in a graph. GNNs operate on graphs and manifolds by generalizing traditional deep learning to irregular domains and thus can deal directly with more general data forms. So far many novel architectures have been put forward and resulted in recent breakthroughs in tasks across various domains \cite{kipf2017semi,ChenL0GZ19,chen2018heterogeneous,ZhangWZ0WL19,BuiYJ19,LeClairHWM20,ying2018graph}. For example, PinSage \cite{ying2018graph}, a GNNs based system, has been successfully deployed at Pinterest\footnote{\url{https://www.pinterest.com/}} recommendation system with millions of users and items.

However, mainstream deep learning frameworks, such as TensorFlow \cite{abadi2016tensorflow} and PyTorch \cite{paszke2019pytorch}, have not yet integrated functional APIs to implement GNNs conveniently and efficiently. Unlike what has been developed and exploited for convolutional neural networks (CNNs), the implementation of core components of GNNs remains a challenge for developers and researchers.
To this end, a line of specialized GNNs toolboxes have been developed for efficient training. For instance, PyG \cite{Fey/Lenssen/2019} and DGL \cite{wang2019dgl} are two popular toolboxes for deep learning on graph-structured data.


While these toolboxes have eased the development of complex GNN models, they also come with steep learning curves for inexperienced developers, since it requires \textit{necessary domain-specific knowledge as well as certain scripting and coding experience in machine learning.}
Developers have sought a platform that can deploy GNN models more conveniently while requiring less expert knowledge. Specifically, there is room for improvement in the following aspects.

\textbf{Deployability}. It is time-consuming, tedious, and demanding for developers who wish to deploy a GNN model. Developers have to build their own pipeline including training and inference procedures from scratch.
\textbf{Reproducibility}.
Developers might not be able to use the existing implementations from published algorithms directly, since they are possibly implemented with various deep learning frameworks. Consequently, developers have to adapt the code of one to another accordingly.
\textbf{Benchmark}. More efforts would be made for researchers who want to conduct benchmarking experiments and performance comparisons. As researchers have to implement different benchmarking models, and
it is necessary but usually tedious work for proper training and fair parameter tuning.


We tackle the aforementioned problems by our proposed GraphGallery, an open-source platform built on TensorFlow and PyTorch. It allows technically inexperienced developers to automatically deploy GNNs based software and also eases the development of new models. In particular, GraphGallery includes the following features:
\begin{itemize}
    \item \textbf{Unified and extensible interface.} As a user-friendly platform, GraphGallery offers unified interfaces for developers to deploy GNNs based software even without expert knowledge.
    \item \textbf{Multiple frameworks support.} GraphGallery interfaces with the most popular deep learning frameworks: TensorFlow and PyTorch.
          Additionally, it integrates seamlessly with PyG, DGL, and other GNNs toolboxes.
    \item \textbf{Fast coding.} GraphGallery provides a model gallery with easy and elegant APIs for developers and researchers, a GNN model can be deployed in a few lines of code.

\end{itemize}

GraphGallery is an out-of-the-box platform and flexibly supports multi-level reuse. Developers with minimum scripting and coding experience are sufficient to deploy a complete GNN model. With this platform, developers can be more focused on designing the architecture of GNNs based software instead of other tedious works. Moreover, GraphGallery provides comprehensive tutorials and examples, which can serve as a good starting point for beginners.


In conclusion, we make the following contributions with our GraphGallery platform:
\begin{itemize}
    \item An extensible, user-friendly, and open-source platform for fast benchmarking and easy development.
    \item A series of building blocks for creating GNNs based software quickly.
    \item A collection of out-of-the-box GNN models for researchers and developers.
\end{itemize}

\section{Related Work}
One of the closest analogs of GraphGallery is PyTorch Geometric (PyG) \cite{Fey/Lenssen/2019} toolbox, which has been recently released as a deep learning library on irregularly structured data. PyG follows a message passing scheme and immutable data flow paradigm and provides the essential building blocks for creating GNNs.



Deep Graph Library (DGL) \cite{wang2019dgl} is another deep learning toolbox developed for graph-structured data. DGL takes the generalized sparse tensor operations as computational schemes of GNNs and advocates graph as the central programming abstraction.



In addition, there are other toolboxes proposed to facilitate the research and development on GNNs, such as NeuGraph \cite{234948} and Spektral \cite{grattarola2020graph}. However, they have been built upon different deep learning frameworks, and the provided APIs are more complicated for inexperienced developers and researchers in this field.

GraphGallery is quite different from the aforementioned toolboxes in both design principles and concepts, it considers more about deploying GNNs based software quickly and easily for developers. We show the comparison of GraphGallery and other toolboxes in Table \ref{tab1}, here we mainly compare the key differences with PyG and DGL. Importantly, GraphGallery is designed initially to lower the bar of entry and accelerate research and development on GNNs based software. To this end, GraphGallery offers unified and easy-to-use APIs for building GNN models and supports various deep learning frameworks. In addition, it provides unified interfaces for more flexible customizations as well as developing interfaces in implementing GNNs.


\begin{table}[t]
    \caption{Comparison of GraphGallery and other GNNs toolboxes.}
    \centering
    \resizebox{\linewidth}{!}{
        \begin{tabular}{@{}lccc@{}}
            \hline
            \textbf{Features}           & \textbf{PyG} & \textbf{DGL} & \textbf{GraphGallery} \\
            \hline
            Built-in Dataset            & \cmark       & \cmark       & \cmark                \\
            Custom Dataset API          & \xmark       & \xmark       & \cmark                \\
            Developing Interface        & \cmark       & \cmark       & \cmark                \\
            Model Gallery               & \xmark       & \xmark       & \cmark                \\
            PyTorch Support             & \cmark       & \cmark       & \cmark                \\
            TensorFlow Support          & \xmark       & \cmark       & \cmark                \\
            Training/Inference Pipeline & \xmark       & \xmark       & \cmark                \\
            \hline
        \end{tabular}
        \label{tab1}
    }
\end{table}

\begin{figure}[t]
    \centering
    \includegraphics[width=\linewidth]{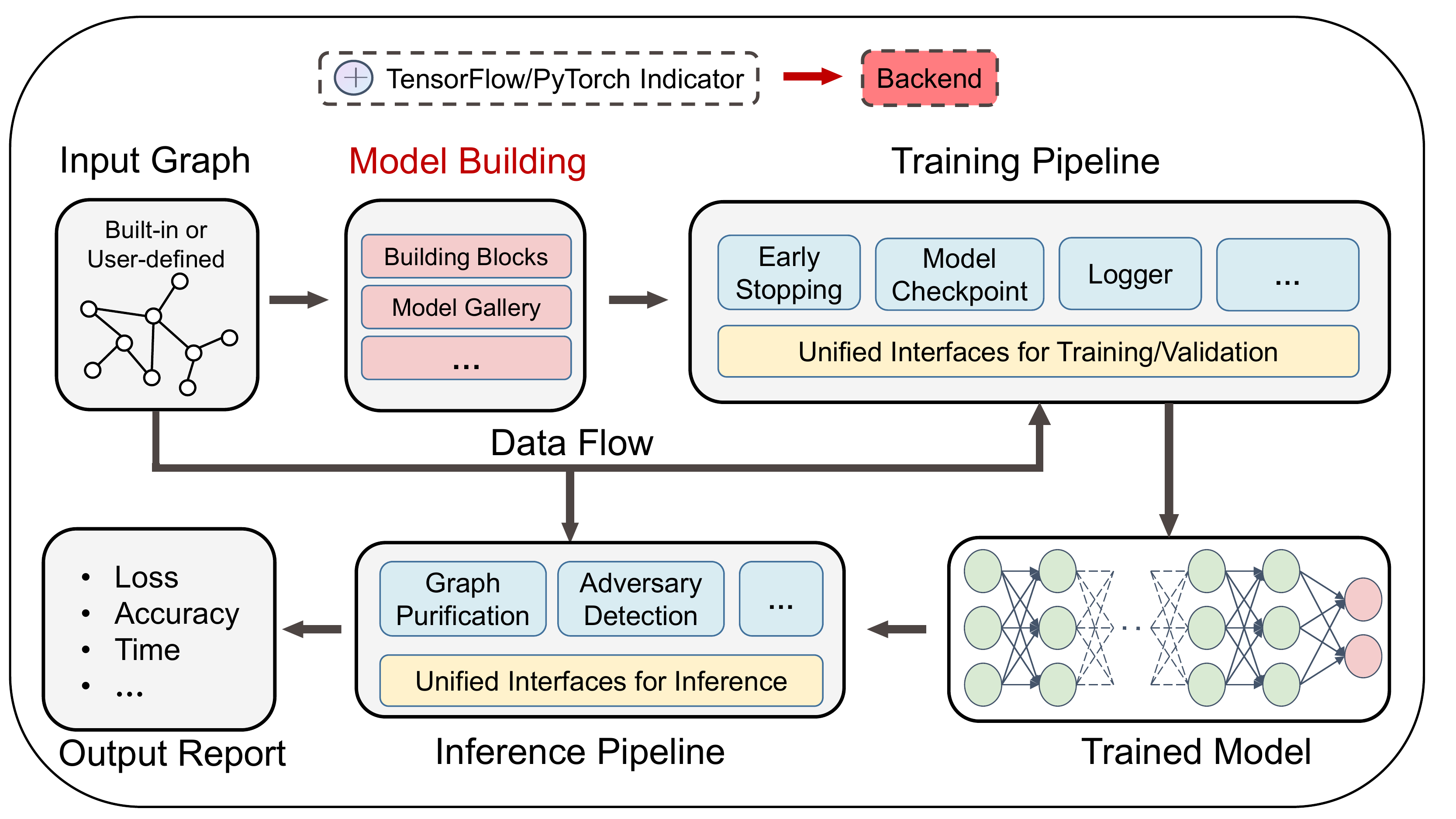}
    \caption{Conceptual architecture of GraphGallery.}
    \label{fig:framework}
\end{figure}

\begin{figure}[t]
    \centering
    \includegraphics[width=.9\linewidth]{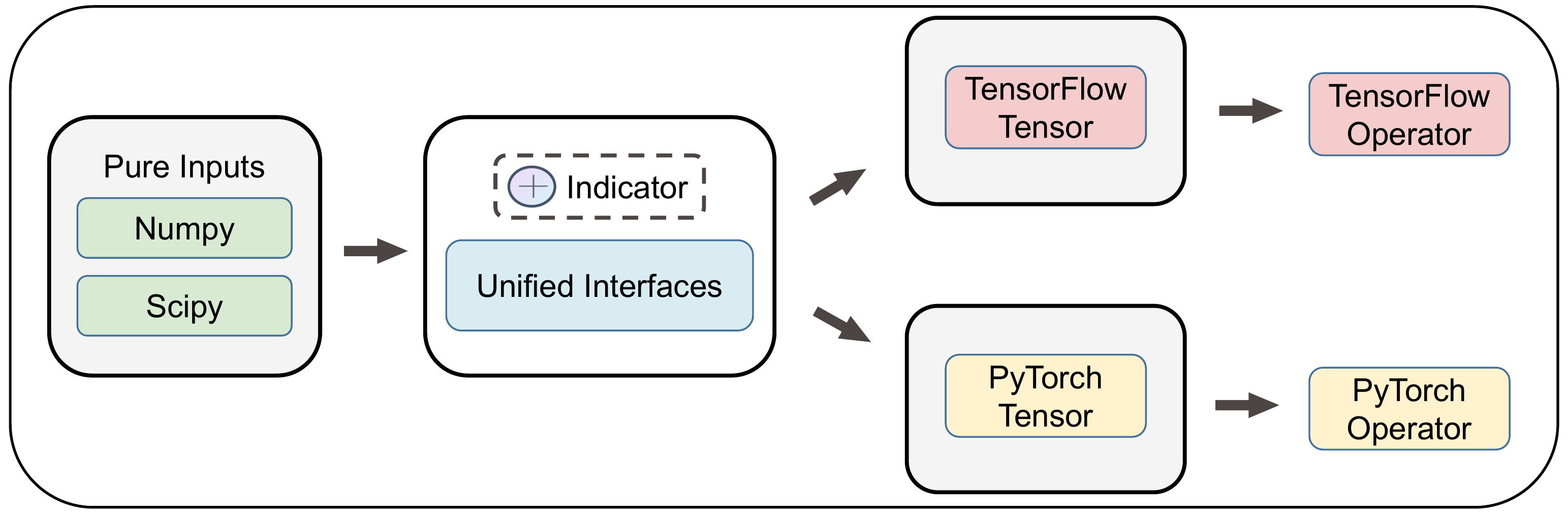}
    \caption{Framework-neutral design of GraphGallery.}
    \label{fig:design}
\end{figure}

\section{GraphGallery Platform}

The overall architecture of GraphGallery is presented in Fig. \ref{fig:framework}. GraphGallery follows an end-to-end design and provides a set of unified interfaces for developers. Such interfaces enable developers to build a custom GNN model and access the intermediate results during training and testing. With these interfaces, developers could be more concerned with the design of the model (``Model Building'' in Fig. \ref{fig:framework}) while other tedious work such as parameter tuning and break-point resume is automatically done by this platform. Besides, we also consider the robustness and stability of GNNs\cite{chen2020survey}, where several specialized methods have been integrated into the inference pipeline for enhancing the robustness of GNNs against adversarial attacks.

GraphGallery has high independility which allows developers to change any components of the framework without impacting the other. For instance, developers could add custom code in the components to redefine the data flow through the model while keeping the rest of the code virtually unchanged. Such a design of GraphGallery endows its maintainability, understandability, and extensibility. Moreover, developers are encouraged to extend GraphGallery to support other GNNs toolboxes (see Sec. \ref{extensibility}).


Similar to the popular DGL toolbox, GraphGallery adopts a framework-neutral design instead of being framework-agnostic. That is, a PyTorch model still needs to be made some adjustments if it is to be run in TensorFlow. Since TensorFlow and PyTorch are two different deep learning frameworks and vary in many aspects, being completely framework-agnostic requires massive effort over all conceivable operators across the two frameworks, and this makes it hard to maintain and extend. Instead, we adopt another practical approach to reduce the dependencies as much as possible and endow the platform a higher degree of flexibility.

Figure \ref{fig:design} demonstrates the framework-neutral design of GraphGallery. We adopt Numpy\footnote{https://numpy.org/} and Scipy\footnote{https://www.scipy.org/} compressed sparse row (CSR) matrices as pure inputs, which are popular forms of data in machine learning. This is quite different from PyG and DGL that have their own data paradigms, and it benefits the extensibility of GraphGallery. Then, the inputs could be transformed into various framework-specific tensors with unified interfaces. The indicator (dotted square) can be switched dynamically to another backend during program running, developers hardly notice the differences of GNNs when working with TensorFlow or PyTorch.



\section{Features of GraphGallery}
GraphGallery is featured in many aspects, including general and user-defined datasets, comprehensive benchmark models, multiple frameworks support, and high extensibility.


\subsection{Built-in and User-defined Dataset}
GraphGallery has already assembled a number of benchmark datasets commonly seen in the literature. The datasets are saved as a Numpy compressed \textit{NpzFile} (A zipped archive of files) and researchers can directly use them in their works without taking into consideration extra preprocessing. Moreover, it is much easier to define and employ custom datasets with the provided interfaces.


\subsection{Model Gallery}
\label{gallery}
GraphGallery is an easy-to-use platform that provides an out-of-the-box model gallery. It is suitable for the three typical scenarios: (i) A beginner with less domain-specific knowledge who wants to understand how GNN works; (ii) A researcher who wants to conduct benchmarking experiments; (iii) A developer who wants to deploy GNNs on existing applications/systems. Specifically, they do not need to consider the API details with different frameworks backend. Instead, they can call a complete GNN implementation from the model gallery directly by just 3-4 lines of code.

\begin{lstlisting}[language=Python, caption=An example to evaluate GCN on a graph using GraphGallery, label=list:example]
# Load data
data = load_data()
# Load GCN from model gallery
from graphgallery.gallery import GCN
model = GCN(data)
model.process(optional_paras).build(optional_paras)
history = model.train(training_set, validation_set)
report = model.test(testing_set)
\end{lstlisting}

We showcase a typical example for GraphGallery in Listing \ref{list:example}, here a simple GCN \cite{kipf2017semi} with default hyperparameters has been deployed and tested. Developers only need to feed the dataset to initialize the corresponding model class (line 5), and optionally specify the hyperparameters into preprocessing and building methods: \texttt{process} and \texttt{build} (line 6). The complete training pipeline is applied to optimize the model and the training details are automatically recorded (line 7). Finally, GraphGallery will assess the model on the testing set and generate the corresponding report (line 8).


\begin{table*}[t]
    \centering
    \caption{Benchmark: Evaluation Results of Selected Models in Model Gallery.}
    \resizebox{0.9\textwidth}{!}{
        \begin{tabular}{c|ccc|ccc|ccc}
            \toprule
            \multirow{2}{*}{}                      &
            \multicolumn{3}{c|}{\textbf{CiteSeer}} & \multicolumn{3}{c|}{\textbf{Cora}} & \multicolumn{3}{c}{\textbf{PubMed}}                                                                                                                                                                                       \\ \midrule \midrule

                                                   & \textbf{GCN}                       & \textbf{SGC}                        & \textbf{GAT}            & \textbf{GCN}            & \textbf{SGC}            & \textbf{GAT}            & \textbf{GCN}            & \textbf{SGC}            & \textbf{GAT}            \\ \midrule
            PyG                                    & 70.76$\pm$0.81                     & 71.96$\pm$0.05                      & 71.55$\pm$1.18          & 81.22$\pm$0.75          & 80.77$\pm$0.58          & 82.72$\pm$1.08          & 78.43$\pm$0.47          & 78.94$\pm$0.22          & 77.34$\pm$0.45          \\ \midrule
            DGL                                    & 70.76$\pm$1.24                     & 71.98$\pm$0.07                      & 71.03$\pm$0.48          & 81.39$\pm$0.44          & 80.70$\pm$0.54          & 82.18$\pm$0.47          & 78.94$\pm$0.62          & 78.88$\pm$0.07          & 77.86$\pm$0.49          \\ \midrule
            PyTorch                                & 70.93$\pm$0.66                     & 71.93$\pm$0.06                      & 71.63$\pm$0.45          & \textbf{81.25$\pm$0.88} & 80.53$\pm$0.48          & \textbf{82.74$\pm$0.58} & \textbf{79.06$\pm$0.34} & 78.99$\pm$0.21          & 77.62$\pm$0.48          \\ \midrule
            TensorFlow                             & \textbf{71.36$\pm$1.04}            & \textbf{72.80$\pm$0.59}             & \textbf{72.19$\pm$0.39} & 80.48$\pm$0.97          & \textbf{81.33$\pm$0.61} & 81.96$\pm$1.02          & 78.91$\pm$0.36          & \textbf{79.24$\pm$0.29} & \textbf{77.91$\pm$0.45} \\ \bottomrule
        \end{tabular}
        \label{benchmark}
    }
\end{table*}

\subsection{Multiple Frameworks Support}
\label{multiple}
GraphGallery supports multiple frameworks: TensorFlow and PyTorch.  Alternatively, developers can write codes with TensorFlow or PyTorch backend without considering the differences of APIs. In addition, the model gallery introduced in Sec. \ref{gallery} has also multiple versions of implementations.  Developers can dynamically switch it to another backend by using \texttt{graphgallery.set\_backend(backend)}, where \textit{backend} could be \textit{TensorFlow} or \textit{PyTorch}. For example, suppose developers want to deploy a GCN model with PyTorch implementation, they can directly use the codes in Listing \ref{list:example} with an extra line of code \texttt{set\_backend("PyTorch")} above, and likewise for deploying model with TensorFlow implementations. This indeed gives the developers a high degree of flexibility and requires less effort in programming with different frameworks.

\subsection{PyG and DGL Integration}
\label{extensibility}
PyG and DGL can be integrated seamlessly into GraphGallery. For researchers that develop GNN models using PyG or DGL, the integration requires a small amount of effort, since all these works have in common that training and inference can be formulated as a unified operation over reusable data flows. Alternatively, GraphGallery also has a line of implementations using PyG and DGL, which researchers are free to use by dynamically switching the backend to another like Sec.\ref{multiple}, e.g., \texttt{set\_backend("PyG")}.

\section{Benchmark and Evaluation}
In this section, we present numerical experiments with the proposed platform. For our experiment, three mainstream GNNs, GCN \cite{kipf2017semi}, SGC \cite{wu2019simplifying} and GAT \cite{velivckovic2017graph} are selected and evaluated on three common citation datasets (CiteSeer, Cora and PubMed \cite{sen2008collective}) with TensorFlow and PyTorch backend, respectively.

\textbf{Setup.} In order to evaluate the correctness and reliability of GraphGallery implementations, we compare all methods with corresponding implementations from other GNNs toolboxes: PyG and DGL. We keep model architectures and hyperparameters the same as the original literature or implementations. Follow this work \cite{kipf2017semi}, we use fixed dataset splits and report the average accuracy and standard deviation for semi-supervised node classification performance over ten runs with different random seeds. The experiments are running on a single NVIDIA TITAN RTX GPU, using TensorFlow 2.1.2, PyTorch 1.6, PyG 1.6.1, and DGL 0.5.2.

\textbf{Reliability of implementation.} Table \ref{benchmark} shows the performance of three GNN models with different implementations. Here we only show partial results of our experiments for the sake of brevity.
As we can see, accuracy values obtained by our PyTorch implementation are almost the same with PyG and DGL (here DGL adopts PyTorch backend). The differences are within 0.5\%. The remaining TensorFlow backend has a relatively larger difference than the rest due to different deep learning frameworks.  However, the maximum difference is 0.8\% and the average difference is admissible around 0.6\%. Therefore, the above results show that our implementations are reliable across different frameworks and toolboxes.

\textbf{Fast coding.} As described in Listing \ref{list:example}, the benchmarking experiments could be finished within a few lines of code. Specifically, after loading the data, the model can be set up automatically. Moreover, training and inference pipelines are built with the provided interfaces. And for developers who wish to use another framework implementation of the selected model, what they simply need is one extra line above to call the \texttt{set\_backend} function. In a word, the whole benchmarking experiment takes much less effort and could be completed within a few lines of code. In contrast, if developers plan to run benchmarks on a couple of models without GraphGallery, they need to implement their own training and inference procedures while considering lots of tedious works, which involves tens of code and more time. Above all, GraphGallery is a fast-coding and user-friendly platform for researchers and developers.

\section{Conclusion}
We open-source GraphGallery, a platform for fast benchmarking and easy development of GNNs based software.  GraphGallery supports multiple frameworks, it hides cumbersome details from developers and performs training and optimization automatically. It is easy for absolute beginners and inexperienced developers to deploy GNNs based software, and it also benefits researchers to simplify implementing and working with GNNs. In the future, we hope more researchers and developers in the deep graph learning community will use GraphGallery for software development and benchmarking experiment to accelerate their works.

\section{Acknowledgement}
The research is supported by the Key-Area Research and Development Program of Guangdong Province (No. 2020B010165003), the National Natural Science Foundation of China (No. U1711267),  the Guangdong Basic and Applied Basic Research Foundation (No. 2020A1515010831), and the Program for Guangdong Introducing Innovative and Entrepreneurial Teams (No. 2017ZT07X355).

\bibliographystyle{IEEEtran}
\bibliography{main}

\end{document}